\documentclass[conference]{IEEEtran}
\IEEEoverridecommandlockouts
\usepackage{cite}
\usepackage{amsmath,amssymb,amsfonts}
\usepackage{graphicx}
\usepackage{float}
\usepackage{multirow}
\usepackage{listings}
\usepackage{multirow} % In your preamble
\usepackage{booktabs}

\usepackage[linesnumbered, ruled, vlined]{algorithm2e}

\def\BibTeX{{\rm B\kern-.05em{\sc i\kern-.025em b}\kern-.08em
    T\kern-.1667em\lower.7ex\hbox{E}\kern-.125emX}}
\begin{document}

\title{CEEMDAN-Based Multiscale CNN for Wind Turbine Gearbox Fault Detection}
\author{
Nejad Alagha$^{*}$, Anis Salwa Mohd Khairuddin$^{\dagger}$, Obada Al-Khatib$^{\ddagger}$, Abigail Copiaco$^{*}$\\
$^*$College of Engineering and Information Technology, University of Dubai, UAE (nalagha@ud.ac.ae;\\
acopiaco@ud.ac.ae)\\

$^\dagger$Department of Electrical Engineering, University of Malaya, Kuala Lumpur, Malaysia (anissalwa@um.edu.my)\\
$^\ddagger$School of Engineering, University of Wollongong in Dubai, Dubai, UAE (obadaalkhatib@uowdubai.ac.ae)
}

\maketitle

\begin{abstract}
Wind turbines play a critical role in the shift toward sustainable energy generation. Their operation relies on multiple interconnected components, and a failure in any of these can compromise the entire system's functionality. Detecting faults accurately is challenging due to the intricate, non-linear, and non-stationary nature of vibration signals, influenced by dynamic loading, environmental variations, and mechanical interactions. As such, effective signal processing techniques are essential for extracting meaningful features to enhance diagnostic accuracy. This study presents a hybrid approach for fault detection in wind turbine gearboxes, combining Complete Ensemble Empirical Mode Decomposition with Adaptive Noise (CEEMDAN) and a Multiscale Convolutional Neural Network (MSCNN). CEEMDAN is employed to decompose vibration signals into intrinsic mode functions, isolating critical features at different time-frequency scales. These are then input into the MSCNN, which performs deep hierarchical feature extraction and classification. The proposed method achieves an F1 Score of 98.95\%, evaluated on real-world datasets, and demonstrates superior performance in both detection accuracy and computational speed compared to existing approaches. This framework offers a balanced solution for reliable and efficient fault diagnosis in wind turbine systems.
\end{abstract}


\section{Introduction}
Renewable energy has emerged as a linchpin in addressing the dual challenges of climate change and sustainable development, playing a vital role in the global energy transition. According to the International Energy Agency (IEA), in 2023, renewable energy sources collectively accounted for 29.6\% of global electricity generation, marking a significant increase from previous years \cite{li2022review}. By 2030, the total greenhouse gas emissions are expected to reach 40 gigatons \cite{keles2012renewable}, proving that the world is facing a major issue in the reduction of greenhouse gas emissions; hence, the importance of renewable energy cannot be overstated, and there is a need to carry out more research to enhance the operation of renewable energy systems. Taking the European countries as an example, in 2017, there was a noticeable decrease in greenhouse emissions, and the reduction even went lower than the 2020 target.

After hydro power, the second most growing renewable energy sector is the wind energy sector, due to its simple infrastructure, cost-effectiveness, and its advanced technology \cite{breeze2014wind}. A wind farm simply is a collection of wind turbines set up either offshore or onshore \cite{hamdaoui2017dynamic} and is connected to the power transmission grid. Although most of the farms currently exist onshore, it is expected that there will be a growth in the construction of offshore wind farms \cite{bilgili2022global}. IEA has identified a noticeable growth in the power generated from offshore wind farms, specifically from the global leaders in offshore wind, such as China, Germany, and the UK \cite{weforum2022offshore}. 

Despite the rapid expansion of wind energy infrastructure, ensuring the operational reliability of wind turbines remains a critical concern. Gearbox faults, in particular, are among the primary causes of unplanned downtime and high maintenance costs \cite{app10093258}. Traditional fault diagnosis methods often fall short when handling the non-stationary and nonlinear nature of vibration signals generated during gearbox operation. This challenge underscores the need for intelligent diagnostic frameworks capable of efficiently processing raw signals and delivering accurate predictions under real-world conditions.

\section{Related Work}
Recent advancements in wind turbine fault diagnosis highlight the growing effectiveness of hybrid methods that combine advanced signal decomposition with intelligent classification models. These approaches are crucial for analyzing complex, nonstationary, and noise-contaminated vibration signals. The Complete Ensemble Empirical Mode Decomposition with Adaptive Noise (CEEMDAN) has gained wide adoption due to its ability to reduce mode mixing, where oscillations of different frequency scales appear within the same Intrinsic Mode Function (IMF) or spread across several IMFs—leading to improved signal reconstruction.

Several studies have applied CEEMDAN to enhance feature extraction in vibration-based fault diagnosis. Zhang \textit{et al.} \cite{9683813} proposed a sliding-window CEEMDAN (SW-CEEMDAN) with a Concept-adapting Very Fast Decision Tree (CVFDT) to address endpoint effects and concept drift in data streams. Energy entropy features from selected IMFs enabled adaptive classification of bearing health states with 95.5\% accuracy using the Case Western Reserve University (CWRU) dataset, demonstrating effective real-time diagnosis.

Li \textit{et al.} \cite{9788916} integrated CEEMDAN with Fast Spectral Kurtosis (FK) for bearing fault identification under strong noise. After IMF selection based on kurtosis and correlation thresholds, reconstructed signals were processed by FK and Hilbert envelope demodulation to extract fault frequencies, successfully enhancing inner-ring defect detection under noisy conditions.

Zhang \textit{et al.} \cite{en17040819} developed a CEEMDAN-CWT Time-Frequency Representation (TFR) method using a kurtosis-correlation (KC) index and Z-threshold for selecting IMFs. Analyzing the selected IMFs in the time-frequency domain enabled identification of inner and outer race bearing faults, outperforming classical demodulation and resonance-based methods in weak signal conditions.

A hybrid CEEMDAN–Grey Wolf Optimized Kernel Extreme Learning Machine (GWO-KELM) model \cite{en16010048} further improved classification accuracy. Following wavelet denoising and CEEMDAN, fuzzy entropy features were optimized via GWO to tune KELM parameters, achieving 99.42\% accuracy on the CWRU dataset, showing the strength of entropy-based features with evolutionary optimization.

Similarly, an Adaptive Chaotic Particle Swarm Optimized Backpropagation (ACPSO-BP) network \cite{articleSS} employed CEEMDAN decomposition and Weighted Permutation Entropy (WPE) for feature extraction. The chaotic PSO improved convergence and discriminability, yielding high accuracy with limited samples and addressing conventional BP training limitations.

Multiscale Convolutional Neural Networks (MSCNNs) can extract hierarchical temporal and spatial features \cite{8384293}. When combined with CEEMDAN, they process IMFs to capture both high- and low-frequency components. Huang \textit{et al.} \cite{9612722} demonstrated the benefit of multiscale CNNs for recognizing structural damage in wind turbine gearboxes. However, few works have applied CEEMDAN-MSCNN frameworks specifically to gearbox fault diagnosis.

Most existing CEEMDAN-based studies focus on bearing faults, while gearbox-level diagnostics remain underexplored. Moreover, most rely on shallow models or handcrafted features rather than end-to-end deep learning. This motivates the proposed CEEMDAN-MSCNN framework, which leverages decomposed vibration signals to train a multiscale deep network for fault classification. Building upon prior decomposition and learning advances, this approach aims to achieve high accuracy, robustness to signal variation, and practical applicability in real-world wind turbine diagnostics addressing the current gap in multiscale deep learning integration with CEEMDAN for gearbox faults.

\section{Methodology}
Based on the gaps identified in the preceding section, the proposed methodology integrates advanced signal processing and deep learning to achieve accurate and robust fault classification in time-series data. Raw signals from multiple sensors, often non-linear and non-stationary, are first decomposed using CEEMDAN into IMFs, each representing a distinct frequency band. These IMFs capture essential multi-scale temporal features for distinguishing between healthy and faulty states. The resulting IMF matrices are fed into a MSCNN, which extracts hierarchical features across spatial (IMF levels) and temporal (sample points) dimensions using convolutional layers. Max-pooling, dropout, and fully connected layers further enhance generalization. 

\subsection{Dataset Description}\label{AA}
This study uses the NREL Gearbox Reliability Collaborative dataset, which focuses on vibration-based monitoring of wind turbine gearboxes \cite{nrel_gearbox_dataset}. It includes data from both healthy and damaged gearboxes tested under controlled conditions, with faults such as scuffing,overheating, and fretting corrosion induced by oil-loss events. The gearbox is a three-stage planetary system. Vibration data were collected from seven accelerometers (AN3–AN7, AN9-AN10) placed at critical points, sampled at 40kHz.
\begin{table}[!t]
\centering
\caption{Summary of Dataset Characteristics}
\label{tab:dataset_summary}
\begin{tabular}{|l|c|}
\hline
\textbf{Attribute} & \textbf{Value} \\
\hline
Sampling Frequency & 40 kHz \\
Sensors Used & 7 (AN3–AN7, AN9-AN10) \\
Signal Type & Vibration (Accelerometer) \\
Conditions & Healthy and Damaged \\
Recordings per Condition & 10 files × 1-minute each \\
Total Duration & 140 minutes (70 Healthy + 70 Damaged) \\
Samples per Sensor File & 2.4 million \\
Total Samples & 336 million \\
\hline
\end{tabular}
\end{table}
The dataset comprises ten 1-minute recordings per condition, offering rich input for fault diagnosis and predictive maintenance. A summarized overview of the dataset content is provided in Table~\ref{tab:dataset_summary}.
\begin{figure*}[ht]
    \centering
    \includegraphics[width=0.95\linewidth]{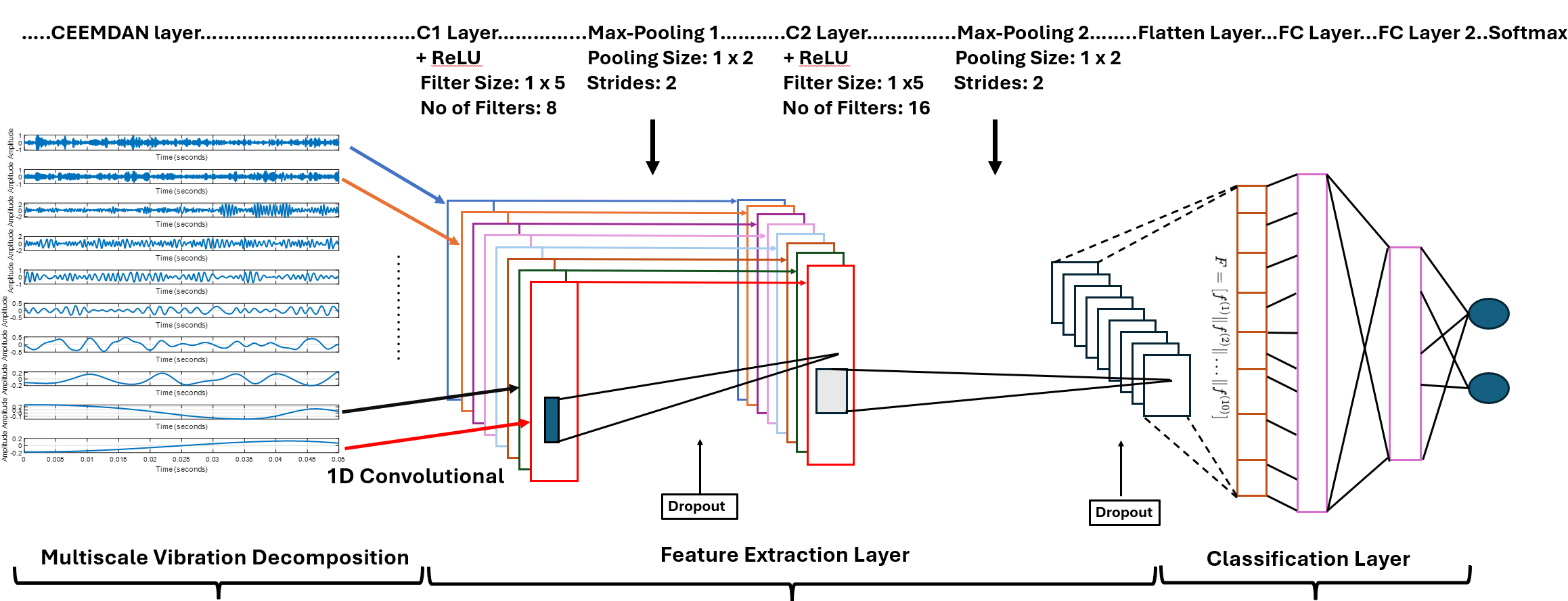}
    \caption{The architecture of the proposed CEEMDAN-based Multiscale CNN model for wind turbine gearbox fault detection.}
    \label{fig:arc}
\end{figure*}

\subsection{Data Pre-processing}
The data was first imported into MATLAB and then consolidated into a single table for structured processing. To increase the number of training samples, each signal was divided into ten non-overlapping windows, with each window containing 20{,}000 samples, equivalent to 0.5 seconds of data. This process was repeated across ten healthy samples and ten damaged samples, resulting in 700 windows for each class. In total, 1400 windows were extracted and labeled, with healthy windows assigned the label 0 and damaged windows the label 1. The final dataset was then randomly shuffled to eliminate any ordering bias.

\subsection{CEEMDAN-MSCNN Approach}
The combination of CEEMDAN and MSCNN offers a powerful framework for wind turbine fault diagnosis, especially when applied to non-stationary and noisy vibration signals like those in our dataset. CEEMDAN decomposes each signal into IMFs, each capturing a specific frequency band, which effectively reveals fault-related patterns that might be obscured in the raw signal. MSCNN complements this by learning hierarchical and scale-specific features from each IMF independently through parallel convolutional branches. This pairing ensures that both localized transient faults (captured in high-frequency IMFs) and long-term structural variations (represented in low-frequency IMFs) are learned simultaneously. 

\subsubsection{Signal Processing using CEEMDAN}
Signal processing transforms raw vibration data into analyzable components for fault classification. This study employs CEEMDAN, a noise-assisted method well suited for non-linear and non-stationary signals. It improves upon EMD and EEMD by introducing adaptive noise to enhance mode separation and reduce sensitivity to signal variability. Unlike EMD, which suffers from mode mixing \cite{xu2019emd}, and EEMD, which may cause reconstruction errors, CEEMDAN iteratively updates residuals, ensuring accurate signal reconstruction \cite{wang2021comparison}.  

CEEMDAN decomposes a signal $x(t)$ into a sum of Intrinsic Mode Functions (IMFs) and a residual:
\begin{equation}
x(t) = \sum_{i=1}^{N_{\text{IMF}}} IMF_i(t) + r(t)
\end{equation}
where each $IMF_i(t)$ captures oscillations at a specific frequency scale and $r(t)$ is the non-oscillatory residual.  

To stabilize decomposition, Gaussian white noise $w(t)$ is added across multiple realizations ($NR$):
\begin{equation}
x_r(t) = x(t) + w_r(t), \quad r = 1,2,\ldots,NR
\end{equation}
Each realization yields IMFs that are averaged to form the final components:
\begin{equation}
IMF_1(t) = \frac{1}{NR} \sum_{r=1}^{NR} IMF_{1,r}(t)
\end{equation}
After each extraction, the residual is updated as:
\begin{equation}
r_k(t) = r_{k-1}(t) - IMF_k(t)
\end{equation}

Key CEEMDAN parameters include the Number of Realizations (NR), Maximum Iterations (MaxIter), and SNR Flag. NR ensures robustness by averaging multiple trials, MaxIter limits decomposition depth to prevent overfitting, and the SNR Flag controls adaptive noise amplitude to separate closely spaced frequencies. Section IV discusses parameter tuning, while Algorithm~1 summarizes the CEEMDAN decomposition steps.

\subsubsection{MSCNN Architecture}
The proposed MSCNN leverages the multiscale frequency characteristics of vibration signals by utilizing the IMFs extracted through CEEMDAN decomposition. Instead of stacking all IMFs into a single matrix, each IMF is processed independently through parallel convolutional branches as shown in Fig.~\ref{fig:arc}, enabling the network to learn discriminative features at distinct frequency bands.

The MSCNN consists of three main parts: (1) multiscale input from CEEMDAN, (2) parallel CNN branches for feature extraction, and (3) a shared classification head.

\textbf{Multiscale Input via CEEMDAN} — Each vibration signal is decomposed into $K=10$ IMFs, $\{IMF_1, IMF_2, \ldots, IMF_{10}\}$, representing different time–frequency scales from high-frequency transients to low-frequency structural patterns. These IMFs are separately input to the CNN, preserving signal diversity without resampling.

\textbf{Parallel Feature Extraction} — Each IMF passes through an identical CNN branch consisting of:
\begin{itemize}
  \item 1D convolution (8 filters, $1\times5$) with ReLU activation;
  \item max-pooling ($1\times2$) and dropout (0.6);
  \item second 1D convolution (16 filters, $1\times5$) with ReLU;
  \item another max-pooling ($1\times2$) and dropout (0.6);
  \item flatten layer producing feature vector $f^{(k)}$.
\end{itemize}
This yields ten feature vectors $f^{(1)}, f^{(2)}, \ldots, f^{(10)}$.

\textbf{Feature Fusion and Classification} — The feature vectors are concatenated:
\[
F = [f^{(1)} \| f^{(2)} \| \ldots \| f^{(10)}]
\]
and passed to a classification head comprising:
\begin{itemize}
  \item fully connected layer (32 neurons, ReLU);
  \item dropout (0.7);
  \item output layer (2 neurons: healthy/faulty) with softmax:
  \[
  P(y=c|F) = \frac{e^{z_c}}{\sum_{j=1}^{2} e^{z_j}}
  \]
\end{itemize}

\textbf{Training and Optimization} — The network is trained using Adam (learning rate $1\times10^{-5}$), batch size 16, and up to 100 epochs. Cross-entropy loss is minimized, and early stopping halts training if validation loss stagnates for 15 epochs.

This multibranch MSCNN enables each frequency component to be analyzed independently, capturing both transient and persistent fault features, and improving generalization compared to conventional single-stream CNNs.

\begin{algorithm}[!t]
\caption{CEEMDAN-Based Signal Decomposition}

\KwIn{Raw vibration signal $x(t)$, number of realizations $NR$, noise amplitude $\epsilon$, maximum number of IMFs $K=10$}
\KwOut{First 10 Intrinsic Mode Functions $\{IMF_1, IMF_2, \ldots, IMF_{10}\}$}

Initialize: $r_0(t) \gets x(t)$ \tcp*{Initial residual is the raw signal}

\For{$k = 2$ \KwTo $10$}{
    \For{$r = 1$ \KwTo $NR$}{
        Generate white noise $w_r(t)$\;
        Add reconstructed signal: 
        $x_r^{(k)}(t) \gets r_{k-1}(t) + \epsilon \cdot \sum_{j=1}^{k-1} IMF_j(t) + w_r(t)$\;
        Apply EMD to $x_r^{(k)}(t)$ to extract first IMF: $IMF_{k,r}(t)$\;
    }
    Compute averaged $k$-th IMF: $IMF_k(t) = \frac{1}{NR} \sum_{r=1}^{NR} IMF_{k,r}(t)$\;
    Update residual: $r_k(t) = r_{k-1}(t) - IMF_k(t)$\;
}

\Return{$\{IMF_1(t), IMF_2(t), \ldots, IMF_{10}(t)\}$}

\end{algorithm}

\begin{table}[!b]
\caption{Validation Accuracy for Different CEEMDAN Parameters}
\vspace{-5pt}
\begin{center}
\begin{tabular}{|c|c|c|c|}
\hline
\textbf{NR} & \textbf{MaxIter} & \textbf{SNRFlag} & \textbf{Validation Accuracy (\%)} \\
\hline
25  & \multirow{4}{*}{250} & \multirow{4}{*}{1} & 80 \\
50  &                      &                    & 90 \\
75  &                      &                    & 90 \\
100 &                      &                    & 85 \\
\hline
\multirow{4}{*}{50} 
     & 100 &\multirow{3}{*}{1}   & 80 \\
     & 250 &              & 90 \\
     & 500 &              & 85 \\
     \hline
50    & 250  & 0             & 80 \\
\hline
\end{tabular}
\label{tab:tuning_results}
\end{center}
\end{table}

\section{Results and Analysis}
In this section, the performance of the MSCNN and the influence of signal processing parameters are examined. The dataset consisted of 1,400 samples, each representing a 0.5-second segment of vibration data from seven sensors under both healthy and damaged gearbox conditions. The data was split into 80\% for training and 20\% for testing. All experiments were conducted on a workstation equipped with an Intel Core i9 CPU, NVIDIA GeForce RTX 4060 GPU, and 32 GB RAM. CEEMDAN parameters including the number of realizations (NR), maximum iterations (MaxIter), and SNR flag were systematically tested to identify the optimal configuration for meaningful IMF extraction. The MSCNN model architecture and hyperparameters were directly adopted from \cite{huang2023wpdmscnn}.

\begin{table}[!t]
\centering
\caption{Impact of Data Length on Model Performance}
\label{tab:data_length}
\begin{tabular}{|c|c|c|}
\hline
\textbf{Data Length (sec)} & \textbf{Accuracy (\%)} & \textbf{F1 Score} \\
\hline
0.10 & 60  & 0.3287 \\
0.15 & 81  & 0.7611 \\
0.20 & 92  & 0.9084 \\
0.25 & 99  & 0.9895 \\
\hline
\end{tabular}
\end{table}

\subsection{Impact of CEEMDAN parameters on training-validation}
To optimize CEEMDAN parameters, we varied the Number of NR, MaxIter, and SNRFlag. The best validation accuracy (90\%) was achieved with $NR = 50, MaxIter = 250,$ and $SNRFlag = 1$. Increasing NR or MaxIter beyond these values led to negligible gains but higher computational cost. Adaptive noise $(SNRFlag = 1)$ significantly improved IMF quality compared to SNRFlag = 0, aligning with prior studies \cite{ wenjun2023denoising}.

\subsection{Algorithm Evaluation Across Different Signal Durations}

\begin{figure}[!b]
    \centering
    \includegraphics[width=0.9\linewidth]{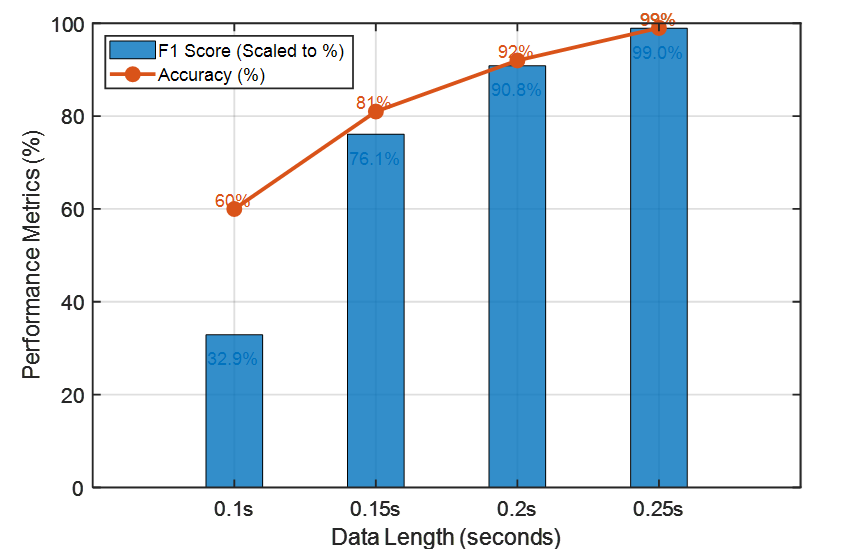}
    \caption{Performance Metrics Across Different Signal Durations}
    \label{fig:duration_accuracy}
\end{figure}

To evaluate the MSCNN model’s accuracy and robustness, we tested it on unseen vibration data with varying signal durations of 0.1, 0.15, 0.2, and 0.25 seconds. This tested the model’s ability to generalize across different input lengths.

As shown in Fig.~\ref{fig:duration_accuracy} and Table~\ref{tab:data_length}, performance improved consistently with longer signals. At 0.1s, the model achieved 60\% accuracy and an F1 score of 0.3287. In contrast, 0.25s inputs yielded 99\% accuracy and an F1 score of 0.9895. The improvement stems from the richer temporal-frequency content of longer signals, which better captures fault-related features. These results demonstrate that 0.25 seconds offers an optimal trade-off between accuracy and computational efficiency for real-world deployments.

\section{Comparative Analysis with Existing Methods}

This section provides a comparative analysis of the proposed method against previous works, the comparison is based on F1 Score, training time, and testing time. The methods compared include MSCNN \cite{jiang2019mscnn}, WPD-MSCNN \cite{huang2023wpdmscnn}, CEEMDAN-BT-CNN \cite{athisayam2024smartceemdan}, TWSVM \cite{dhiman2021tws} and the proposed method, highlighting differences in accuracy, computational efficiency, and inference performance.

As shown in Table~\ref{tab:comparison}, the proposed method achieves an F1 Score of 98.95\%, demonstrating competitive accuracy, surpassing MSCNN (98.53\%) and CEEMDAN-BT-CNN (97.5\%) while slightly trailing behind WPD-MSCNN (99.43\%). In terms of computational efficiency, the proposed method excels with a training time of 2.26 seconds per epoch over 100 epochs, significantly faster than MSCNN’s 14.54 seconds per epoch and comparable to WPD-MSCNN’s 3.63 seconds per epoch. Overall, the proposed method strikes a balance between high classification performance and computational efficiency.
\begin{table}[t]
\caption{Performance Comparison with Recent Works}
\centering
\scriptsize % Smaller font to ensure it fits the column
\renewcommand{\arraystretch}{1.5} % Improves row spacing
\begin{tabular}{|l|c|c|c|}
\hline
\textbf{Previous Works} & \textbf{F1 (\%)} & \textbf{Train Time} & \textbf{Test Time} \\
 & & \textbf{(s/epoch)} & \textbf{(ms/sample)} \\
\hline
MS-CNN \cite{jiang2019mscnn}         & 98.53 & 14.54 (50)    & 0.18 \\
CEEMDAN-BT-CNN \cite{athisayam2024smartceemdan} & 97.5  & N/A            & N/A \\
WPD-MSCNN \cite{huang2023wpdmscnn}       & 99.43 & 3.63 (100)     & 0.17 \\
TWSVM \cite{dhiman2021tws}         & 95.84 & 1.63 (100)     & N/A \\
\textbf{Proposed Method}      & \textbf{98.95} & \textbf{2.26 (100)} & \textbf{0.28} \\
\hline
\end{tabular}
\label{tab:comparison}
\end{table}
\section{Conclusion}

This study presented a hybrid fault diagnosis framework combining CEEMDAN for signal decomposition and a MSCNN for classification of wind turbine gearbox faults. Through structured parameter tuning, the CEEMDAN configuration was optimized to enhance IMF quality and model performance. The proposed method demonstrated strong generalization across varying signal durations, maintaining high accuracy and F1 Score even with short input segments.

Comparative analysis with recent approaches highlighted the method’s competitive accuracy, as well as superior computational efficiency, achieving significantly reduced training and testing times. The model can be implemented on edge devices and integrated with existing SCADA or condition monitoring infrastructure.

Future work may explore extending the model to multi-class fault scenarios and validating performance in real-world operational environments.



\begin{thebibliography}{10}
\providecommand{\url}[1]{#1}
\csname url@samestyle\endcsname
\providecommand{\newblock}{\relax}
\providecommand{\bibinfo}[2]{#2}
\providecommand{\BIBentrySTDinterwordspacing}{\spaceskip=0pt\relax}
\providecommand{\BIBentryALTinterwordstretchfactor}{4}
\providecommand{\BIBentryALTinterwordspacing}{\spaceskip=\fontdimen2\font plus
\BIBentryALTinterwordstretchfactor\fontdimen3\font minus \fontdimen4\font\relax}
\providecommand{\BIBforeignlanguage}[2]{{%
\expandafter\ifx\csname l@#1\endcsname\relax
\typeout{** WARNING: IEEEtran.bst: No hyphenation pattern has been}%
\typeout{** loaded for the language `#1'. Using the pattern for}%
\typeout{** the default language instead.}%
\else
\language=\csname l@#1\endcsname
\fi
#2}}
\providecommand{\BIBdecl}{\relax}
\BIBdecl

\bibitem{li2022review}
L.~Li \emph{et~al.}, ``Review and outlook on the international renewable energy development,'' \emph{Energy and Built Environment}, vol.~3, no.~2, pp. 139--157, Apr. 2022.

\bibitem{keles2012renewable}
S.~Keleş and S.~Bilgen, ``Renewable energy sources in turkey for climate change mitigation and energy sustainability,'' \emph{Renewable and Sustainable Energy Reviews}, vol.~16, no.~7, pp. 5199--5206, Sep. 2012.

\bibitem{breeze2014wind}
P.~Breeze, ``Wind power,'' in \emph{Power Generation Technologies}.\hskip 1em plus 0.5em minus 0.4em\relax Elsevier, 2014, pp. 223--242.

\bibitem{hamdaoui2017dynamic}
Y.~Hamdaoui and A.~Maach, ``Dynamic balancing of powers in islanded microgrid using distributed energy resources and prosumers for efficient energy management,'' in \emph{2017 IEEE International Conference on Smart Energy Grid Engineering (SEGE)}.\hskip 1em plus 0.5em minus 0.4em\relax Oshawa, ON, Canada: IEEE, Aug. 2017, pp. 155--161.

\bibitem{bilgili2022global}
M.~Bilgili and H.~Alphan, ``Global growth in offshore wind turbine technology,'' \emph{Clean Techn Environ Policy}, vol.~24, no.~7, pp. 2215--2227, Sep. 2022.

\bibitem{weforum2022offshore}
{World Economic Forum}, ``Explainer: What is offshore wind and what does its future look like?'' \url{https://www.weforum.org/agenda/2022/11/offshore-wind-farms-future-renewables/}, 2023, accessed: Dec. 09, 2023.

\bibitem{app10093258}
\BIBentryALTinterwordspacing
Z.~Wu, X.~Wang, and B.~Jiang, ``Fault diagnosis for wind turbines based on relieff and extreme gradient boosting,'' \emph{Applied Sciences}, vol.~10, no.~9, 2020. [Online]. Available: \url{https://www.mdpi.com/2076-3417/10/9/3258}
\BIBentrySTDinterwordspacing

\bibitem{9683813}
F.~Shi, J.~Yu, M.~Gu, K.~Lei, and J.~He, ``Research on wind turbine gearbox fault diagnosis based on ceemdan and cvfdt,'' in \emph{2021 11th International Conference on Power and Energy Systems (ICPES)}, 2021, pp. 713--717.

\bibitem{9788916}
X.~Li, L.~Fu, and G.~Zhu, ``Fault diagnosis of rolling bearing based on ceemdan reconstruction and fast spectral kurtosis,'' in \emph{MEMAT 2022; 2nd International Conference on Mechanical Engineering, Intelligent Manufacturing and Automation Technology}, 2022, pp. 1--5.

\bibitem{en17040819}
\BIBentryALTinterwordspacing
D.~Zhang, Y.~Wang, Y.~Jiang, T.~Zhao, H.~Xu, P.~Qian, and C.~Li, ``A novel wind turbine rolling element bearing fault diagnosis method based on ceemdan and improved tfr demodulation analysis,'' \emph{Energies}, vol.~17, no.~4, 2024. [Online]. Available: \url{https://www.mdpi.com/1996-1073/17/4/819}
\BIBentrySTDinterwordspacing

\bibitem{en16010048}
\BIBentryALTinterwordspacing
L.~Liu, Y.~Wei, X.~Song, and L.~Zhang, ``Fault diagnosis of wind turbine bearings based on ceemdan-gwo-kelm,'' \emph{Energies}, vol.~16, no.~1, 2023. [Online]. Available: \url{https://www.mdpi.com/1996-1073/16/1/48}
\BIBentrySTDinterwordspacing

\bibitem{articleSS}
S.~Song, S.~Zhang, W.~Dong, X.~Zhang, and W.~Ma, ``A new hybrid method for bearing fault diagnosis based on ceemdan and acpso-bp neural network,'' \emph{Journal of Mechanical Science and Technology}, vol.~37, 11 2023.

\bibitem{8384293}
G.~Jiang, H.~He, J.~Yan, and P.~Xie, ``Multiscale convolutional neural networks for fault diagnosis of wind turbine gearbox,'' \emph{IEEE Transactions on Industrial Electronics}, vol.~66, no.~4, pp. 3196--3207, 2019.

\bibitem{9612722}
D.~Huang, W.-A. Zhang, F.~Guo, W.~Liu, and X.~Shi, ``Wavelet packet decomposition-based multiscale cnn for fault diagnosis of wind turbine gearbox,'' \emph{IEEE Transactions on Cybernetics}, vol.~53, no.~1, pp. 443--453, 2023.

\bibitem{nrel_gearbox_dataset}
{National Renewable Energy Laboratory}, ``{Wind Turbine Gearbox Condition Monitoring Vibration Analysis Benchmarking Datasets},'' \url{https://catalog.data.gov/dataset/wind-turbine-gearbox-condition-monitoring-vibration-analysis-benchmarking-datasets}, 2014, accessed: 2025-06-09.

\bibitem{xu2019emd}
B.~Xu, Y.~Sheng, P.~Li, Q.~Cheng, and J.~Wu, ``Causes and classification of emd mode mixing,'' \emph{Vibroengineering Procedia}, vol.~22, pp. 158--164, Mar. 2019.

\bibitem{wang2021comparison}
S.~Wang, Y.~Shao, J.~Qian, S.~Sun, and S.~Yu, ``The comparison of some algorithm based on ceemdan,'' in \emph{2021 2nd International Conference on Big Data \& Artificial Intelligence \& Software Engineering (ICBASE)}.\hskip 1em plus 0.5em minus 0.4em\relax Zhuhai, China: IEEE, Sep. 2021, pp. 210--217.

\bibitem{huang2023wpdmscnn}
D.~Huang, W.-A. Zhang, F.~Guo, W.~Liu, and X.~Shi, ``Wavelet packet decomposition-based multiscale cnn for fault diagnosis of wind turbine gearbox,'' \emph{IEEE Transactions on Cybernetics}, vol.~53, no.~1, pp. 443--453, Jan. 2023.

\bibitem{wenjun2023denoising}
B.~Wenjun and C.~Yingjie, ``Denoising of blasting vibration signals based on ceemdan-ica algorithm,'' \emph{Scientific Reports}, vol.~13, no.~1, p. 20928, Nov. 2023.

\bibitem{jiang2019mscnn}
G.~Jiang, H.~He, J.~Yan, and P.~Xie, ``Multiscale convolutional neural networks for fault diagnosis of wind turbine gearbox,'' \emph{IEEE Transactions on Industrial Electronics}, vol.~66, no.~4, pp. 3196--3207, Apr. 2019.

\bibitem{athisayam2024smartceemdan}
A.~Athisayam and M.~Kondal, ``A smart ceemdan, bessel transform and cnn-based scheme for compound gear-bearing fault diagnosis,'' \emph{Journal of Vibration Engineering \& Technologies}, vol.~12, no.~S1, pp. 393--412, Dec. 2024.

\bibitem{dhiman2021tws}
H.~Dhiman, D.~Deb, S.~M. Muyeen, and I.~Kamwa, ``Wind turbine gearbox anomaly detection based on adaptive threshold and twin support vector machines,'' \emph{IEEE Transactions on Energy Conversion}, vol.~36, no.~4, pp. 3462--3469, Dec. 2021.

\end{thebibliography}
\end{document}